\documentclass[11pt,a4paper]{article}
\usepackage[hyperref]{acl2021}
\usepackage{times}
\usepackage{latexsym}

\usepackage{microtype}

\usepackage{amsmath}
\usepackage{xcolor}
\usepackage{array}

\usepackage{graphicx}
\usepackage{algorithm}
\usepackage{algorithmic}
\usepackage{booktabs}
\usepackage{multirow}
\usepackage{amsfonts}
\usepackage{amssymb}

\usepackage{nicefrac}
\usepackage{url}

\makeatletter
\newcommand{\ssymbol}[1]{$^{\@fnsymbol{#1}}$}
\makeatother

\newlength\savewidth
\newcommand\shline{\noalign{\global\savewidth\arrayrulewidth
                            \global\arrayrulewidth 0.8pt}
                   \hline
                   \noalign{\global\arrayrulewidth\savewidth}}

\aclfinalcopy

\title{Competence-based Multimodal Curriculum Learning for \\ Medical Report Generation}

\author{Fenglin Liu\textsuperscript{1}, Shen Ge\textsuperscript{2}, Yuexian Zou\textsuperscript{1}, Xian Wu\textsuperscript{2}\\
\textsuperscript{1}School of ECE, Peking University\\
\textsuperscript{2}Tencent Medical AI Lab, Beijing, China\\
{\tt \{fenglinliu98, zouyx\}@pku.edu.cn; \{shenge, kevinxwu\}@tencent.com}\\ 
}

\date{}

\begin{document}
\maketitle
\begin{abstract}

Medical report generation task, which targets to produce long and coherent descriptions of medical images, has attracted growing research interests recently. Different from the general image captioning tasks, medical report generation is more challenging for data-driven neural models. This is mainly due to 1) the serious data bias and 2) the limited medical data. To alleviate the data bias and make best use of available data, we propose a Competence-based Multimodal Curriculum Learning framework (CMCL). Specifically, CMCL simulates the learning process of radiologists and optimizes the model in a step by step manner. Firstly, CMCL estimates the difficulty of each training instance and evaluates the competence of current model; Secondly, CMCL selects the most suitable batch of training instances considering current model competence. By iterating above two steps, CMCL can gradually improve the model's performance. The experiments on the public IU-Xray and MIMIC-CXR datasets show that CMCL can be incorporated into existing models to improve their performance.

\end{abstract}

\begin{figure}[t]

\centering
\includegraphics[width=1\linewidth]{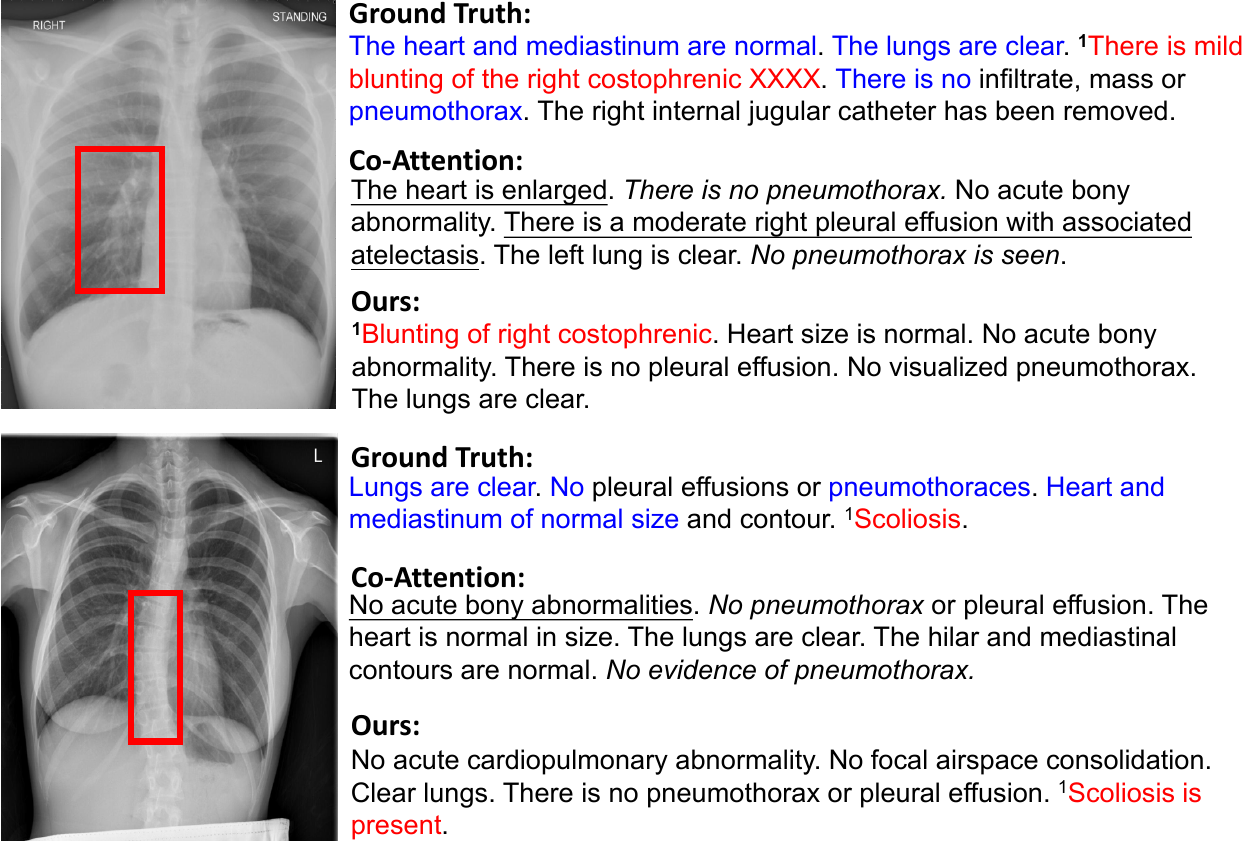}
\caption{Two examples of ground truth reports and reports generated by a state-of-the-art approach Co-Attention \cite{Jing2018Automatic} and our approach. The {\color{red} Red} bounding boxes and {\color{red} Red} colored text indicate the abnormalities in images and reports, respectively. The {\color{blue} Blue} colored text stands for the similar sentences used to describe the normalities in ground truth reports. There are notable visual and textual data biases and the Co-Attention \cite{Jing2018Automatic} fails to depict the rare but important abnormalities and generates some error sentences (\underline{Underlined} text) and repeated sentences (\textit{Italic} text).}
\label{fig:introduction}
\end{figure}

\section{Introduction}
Medical images, e.g., radiology and pathology images, and their corresponding reports, which describe the observations in details of both normal and abnormal regions, are widely-used for diagnosis and treatment \cite{Delrue2011Difficulties,goergen2013evidence}.
In clinical practice, writing a medical report can be time-consuming and tedious for experienced radiologists, and error-prone for inexperienced radiologists.
Therefore, automatically generating medical reports can assist radiologists in clinical decision-making and emerge as a prominent attractive research direction in both artificial intelligence and clinical medicine~\cite{Jing2018Automatic,Jing2019Show,Li2018Hybrid,Li2019Knowledge,Wang2018TieNet,Xue2018Multimodal,Yuan2019Enrichment,Zhang2020When,Chen2020Generating,fenglin2021PPKED,liu2021Contrastive,Liu2019Clinically}.

Many existing medical report generation models adopt the standard image captioning approaches: a CNN-based image encoder followed by a LSTM-based report decoder, e.g., CNN-HLSTM \cite{Jing2018Automatic,Liang2017Hierarchical}.
However, directly applying image captioning approaches to medical images has the following problems:
1) \textbf{Visual data bias}: the normal images dominate the dataset over the abnormal ones \cite{Shin2016Learning}. Furthermore, for each abnormal image, the normal regions dominate the image over the abnormal ones. As shown in Figure~\ref{fig:introduction}, abnormal regions (Red bounding boxes) only occupy a small part of the entire image;
2) \textbf{Textual data bias}: as shown in Figure~\ref{fig:introduction}, in a medical report, radiologists tend to describe all the items in an image, making the descriptions of normal regions dominate the entire report. 
Besides, many similar sentences are used to describe the same normal regions.
3) \textbf{Training efficiency}: during training, most existing works treat all the samples equally without considering their difficulties. As a result, the visual and textual biases could mislead the model training~\cite{Jing2019Show,Xue2018Multimodal,Yuan2019Enrichment,fenglin2021PPKED,liu2021Contrastive,Li2018Hybrid}. As shown in Figure~\ref{fig:introduction}, even a state-of-the-art model \cite{Jing2018Automatic} still generates some repeated sentences of normalities and fails to depict the rare but important abnormalities.

To this end, we propose a novel Competence-based Multimodal Curriculum Learning framework (CMCL) which progressively learns medical reports following an easy-to-hard fashion. Such a step by step process is similar to the learning curve of radiologists: (1) first start from simple and easy-written reports; (2) and then attempt to consume harder reports, which consist of rare and diverse abnormalities.
In order to model the above gradual working patterns, CMCL first assesses the difficulty of each training instance from multiple perspectives (i.e., the Visual Complexity and Textual Complexity) and then automatically selects the most rewarding training samples according to the current competence of the model.
In this way, once the easy and simple samples are well-learned, CMCL increases the chance of learning difficult and complex samples,
preventing the models from getting stuck in bad local optima\footnote{Current models tend to generate plausible general reports with no prominent abnormal narratives \cite{Jing2019Show,Li2018Hybrid,Yuan2019Enrichment,fenglin2021PPKED,liu2021Contrastive}}, which is obviously a better solution than the common approaches of uniformly sampling training examples from the limited medical data.
As a result, CMCL could better utilize the limited medical data to alleviate the data bias. We evaluate the effectiveness of the proposed CMCL on IU-Xray \cite{Dina2016IU-Xray} and MIMIC-CXR \cite{Johnson2019MIMIC}.

Overall, the main contributions of this work are:
\begin{itemize}
    \item We introduce the curriculum learning in medical report generation, which enables the models to gradually proceed from easy samples to more complex ones in training, helping existing models better utilize the limited medical data to alleviate the data bias.
    
    \item We assess the difficulty of each training instance from multiple perspectives and propose a competence-based multimodal curriculum learning framework (CMCL) to consider multiple difficulties simultaneously.
    
    \item We evaluate our proposed approach on two public datasets. After equipping our proposed CMCL, which doesn't introduce additional parameters and only requires a small modification to the training data pipelines, performances of the existing baseline models can be improved on most metrics. Moreover, we conduct human evaluations to measure the effectiveness in terms of its usefulness for clinical practice.
    
\end{itemize}

\section{Related Work}
\label{sec:related}

\subsection{Image Captioning and Paragraph Generation}

The task of image captioning \cite{chen2015microsoft,Vinyals2015Show}, which aims to generate a sentence to describe the given image, has received extensive research interests \cite{rennie2017self,anderson2018bottom,liu2020prophet}. These approaches mainly adopt the encoder-decoder framework which translates the image to a \textit{single} descriptive sentence. Such an encoder-decoder framework have achieved great success in advancing the state-of-the-arts \cite{Vinyals2015Show,lu2017knowing,Xu2015Show,Cornia2020M2,Pan2020XLinear,liu2020prophet}.
Specifically, the encoder network \cite{Krizhevsky2012CNN,he2016deep} computes visual representations for the visual contents and the decoder network \cite{Hochreiter1997LSTM,Vaswani2017Transformer} generates a target sentence based on the visual representations.
In contrast to the image captioning, image paragraph generation, which aims to produce a long and semantic-coherent paragraph to describe the input image, has recently attracted growing research interests \cite{Krause2017Hierarchical,Liang2017Hierarchical,Yu2016Hierarchical}.
To perform the image paragraph generation, a hierarchical LSTM (HLSTM) \cite{Krause2017Hierarchical,Liang2017Hierarchical} is proposed as the decoder to well generate long paragraphs.

\subsection{Medical Report Generation}
The medical reports are expected to 1) cover contents of key medical findings such as heart size, lung opacity, and bone structure; 2) correctly capture any abnormalities and support with details such as the location and shape of the abnormality; 3) correctly describe potential diseases such as effusion, pneumothorax and consolidation \cite{Delrue2011Difficulties,goergen2013evidence,Li2018Hybrid,fenglin2021PPKED,you2021align}.
Therefore, correctly describing the abnormalities become the most urgent goal and the core value of this task.
Similar to image paragraph generation, most existing medical report generation works \cite{Jing2018Automatic,Jing2019Show,Li2018Hybrid,Wang2018TieNet,Xue2018Multimodal,Yuan2019Enrichment,Zhang2020When,Zhang2020Optimizing,Miura2021Factual,Lovelace2020Learning,liu2021Contrastive,Liu2019Clinically} attempt to adopt a CNN-HLSTM based model to automatically generate a fluent report.
However, due to the data bias and the limited medical data, these models are biased towards generating plausible but general reports without prominent abnormal narratives \cite{Jing2019Show,Li2018Hybrid,Yuan2019Enrichment,fenglin2021PPKED,liu2021Contrastive,liu2021KGAE}.

\subsection{Curriculum Learning}
In recent years, curriculum learning \cite{Bengio2009Curriculum}, which enables the models to gradually proceed from easy samples to more complex ones in training \cite{elman1993learning}, has received growing research interests in natural language processing field, e.g., neural machine translation \cite{Platanios2019Competence,Kumar2019Reinforcement,Zhao2020Reinforced,Liu2020Norm,Zhang2018Empirical,Kocmi2017Curriculum,Xu2020Dynamic} and computer vision field, e.g., image classification \cite{Weinshall2018classification}, human
attribute analysis\cite{Wang2019human} and visual question answering \cite{Li2020VQACL}.
For example, in neural machine translation, \citet{Platanios2019Competence} proposed to utilize the training samples in order of easy-to-hard and to describe the ``difficulty'' of a training sample using the sentence length or the rarity of the words appearing in it \cite{Zhao2020Reinforced}.
However, these methods \cite{Platanios2019Competence,Liu2020Norm,Xu2020Dynamic} are single difficulty-based and unimodal curriculum learning approaches. It is obviously not applicable to medical report generation task, which involves multi-modal data, i.e., visual medical images and textual reports, resulting in multi-modal complexities, i.e., the visual complexity and the textual complexity. Therefore, it is hard to design one single metric to estimate the overall difficulty of medical report generation. To this end, based on the work of \citet{Platanios2019Competence}, we propose a competence-based multimodal curriculum learning approach with multiple difficulty metrics.

\begin{figure*}[t]
\centering
\includegraphics[width=1\linewidth]{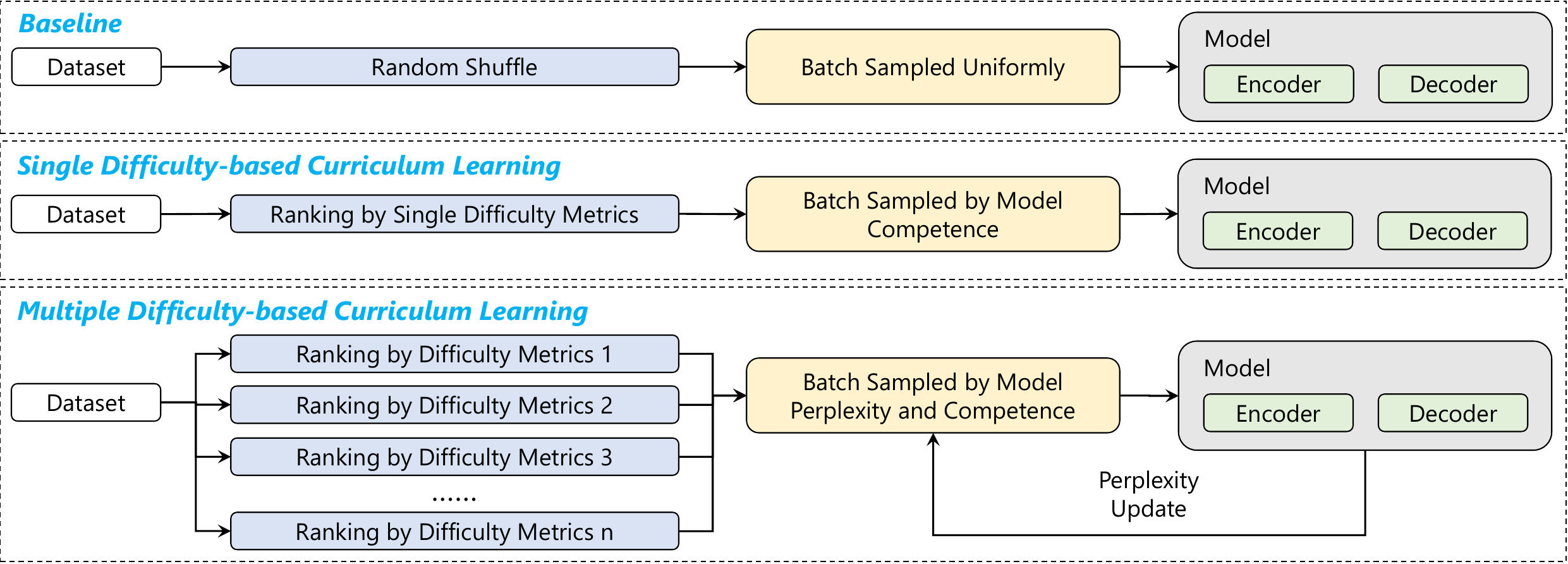}
\caption{The top illustrates the typical encoder-decoder approach; The middle illustrates the Single Difficulty-based Curriculum Learning, where only one difficulty metric is used; The bottom illustrates the  Multiple Difficulty-based Curriculum Learning, where  multiple difficulty metrics are introduced.}
\label{fig:framework}
\end{figure*}

\section{Framework}
\label{sec:framework}

In this section, we briefly describe typical medical report generation approaches and introduce the proposed Competence-based Multimodal Curriculum Learning (CMCL).

As shown in the top of Figure \ref{fig:framework}, many medical report generation models adopt the encoder-decoder manner. Firstly, the visual features are extracted from the input medical image via a CNN model. Then the visual features are fed into a sequence generation model, like LSTM to produce the medical report. In the training phase, all training instances are randomly shuffled and grouped into batches for training. In other words, all training instances are treated equally. Different from typical medical report generation models, CMCL builds the training batch in a selective manner. The middle part of Figure \ref{fig:framework} displays the framework of CMCL equipped with one single difficulty metric. CMCL first ranks all training instances according to this difficulty metric and then gradually enlarges the range of training instances that the batch is selected. In this manner, CMCL can train the models from easy to difficult instances. 

Since medical report generation involves multi-modal data, like visual medical images and textual reports, it is hard to design one single metric to estimate the overall difficulty. Therefore, we also propose a CMCL with multiple difficulty metrics. As shown in the bottom of Figure \ref{fig:framework}, the training instances are ranked by multiple metrics independently. At each step, CMCL generates one batch for each difficulty metric and then calculates the perplexity of each batch based on current model. The batch with highest perplexity is selected to train the model. It can be understood that CMCL sets multiple syllabus in parallel, and the model is optimized towards the one with lowest competence.

\section{Difficulty Metrics}
\label{sec:difficulty}
In this section, we define the difficulty metrics used by CMCL.
As stated in Section~\ref{sec:related}, the key challenge of medical report generation is to accurately capture and describe the abnormalities \cite{Delrue2011Difficulties,goergen2013evidence,Li2018Hybrid}.
Therefore, we assess the difficulty of instances based on the difficulty of accurately capturing and describing the abnormalities.

\subsection{Visual Difficulty}

We define both a heuristic metric and a model-based metric to estimate the visual difficulty.

\paragraph{Heuristic Metric $d_1$}
If a medical image contains complex visual contents, it is more likely to contain more abnormalities, which increases the difficulty to accurately capture them.
To measure such visual difficulty, we adopt the widely-used ResNet-50 \cite{he2016deep} pre-trained on ImageNet \cite{Deng2009ImageNet} and fine-tuned on CheXpert dataset \cite{Irvin2019CheXpert}, which consists of 224,316 X-ray images with each image labeled with occurrences of 14 common radiographic observations.
Specifically, we first extract the normal image embeddings of all normal training images from the last average pooling layer of ResNet-50.
Then, given an input image, we again use the ResNet-50 to obtain the image embedding. 
At last, the average cosine similarity between the input image and normal images is adopted as the heuristic metric of visual difficulty.

\paragraph{Model Confidence $d_2$}
We also introduce a model-based metric.
We adopt the above ResNet-50 to conduct the abnormality classification task.
We first adopt the ResNet-50 to acquire the classification probability distribution $P(I) = \{p_1(I), p_2(I), \dots, p_{14}(I)\}$ among the 14 common diseases for each image $I$ in the training dataset, where $p_n(I) \in [0,1]$.
Then, we employ the entropy value $H(I)$ of the probability distribution, defined as follows:
\begin{equation}
\footnotesize
\begin{aligned}
H(I)=-\sum_{n=1}^{14} &(p_n(I) \log \left(p_n(I)\right) + \\
& \left(1-p_n(I)\right) \log \left(1-p_n(I)\right))
\end{aligned}
\end{equation}
We employ the entropy value $H(I)$ as the model confidence measure, indicating whether an image is easy to be classified or not.

\subsection{Textual Difficulty}
We also define a heuristic metric and a model-based metric to estimate the textual difficulty.

\paragraph{Heuristic Metric $d_3$}
A serious problem for medical report generation models is the tendency to generate plausible general reports with no prominent abnormal narratives \cite{Jing2019Show,Li2018Hybrid,Yuan2019Enrichment}.
The normal sentences are easy to learn, but are less informative, while most abnormal sentences, consisting of more rare and diverse abnormalities, are relatively more difficult to learn, especially at the initial learning stage.
To this end, we adopt the number of abnormal sentences in a report to define the difficulty of a report.
Following \citet{Jing2018Automatic}, we consider sentences which contain ``no'', ``normal'', ``clear'', ``stable'' as normal sentences, the rest sentences are consider as abnormal sentences.

\paragraph{Model Confidence $d_4$}
Similar to visual difficulty, we further introduce a model confidence as a metric.
To this end, we define the difficulty using the negative log-likelihood loss values \cite{Xu2020Dynamic,Zhang2018Empirical} of training samples.
To acquire the negative log-likelihood loss values, we adopt the widely-used and classic CNN-HLSTM \cite{Jing2018Automatic}, in which the CNN is implemented with ResNet-50, trained on the downstream dataset used for evaluation with a cross-entropy loss.

It is worth noting that since we focus on the medical report generation and design the metrics based on the difficulty of accurately capturing and describing the abnormalities, we do not consider some language difficulty metrics used in neural machine translation, e.g., the sentence length \cite{Platanios2019Competence}, the n-gram rarity together with Named Entity Recognition (NER) and Parts of Speech (POS) taggings \cite{Zhao2020Reinforced}.

\section{Approach}
\label{sec:approach}
In this section, we first briefly introduce the conventional single difficulty-based curriculum \cite{Platanios2019Competence}. Then we propose the multiple difficulty-based curriculum learning for medical report generation.

\begin{algorithm}[t]
\small
	\renewcommand{\algorithmicrequire}{\textbf{Input:}}
	\renewcommand{\algorithmicensure}{\textbf{Output:}}
	\caption{\small{Single Difficulty-based Curriculum Learning \cite{Platanios2019Competence}.}}
	\label{alg:single}
	\begin{algorithmic}[1]
		\REQUIRE The training set $D^\text{train}$.
		\ENSURE A model with single difficulty-based curriculum learning.
		\STATE Compute difficulty $d$ for each training sample in $D^\text{train}$;
		\STATE Sort $D^\text{train}$ based $d$ to acquire $D^\text{train}_1$;
		\STATE At $t = 0$, initialize the model competence $c(0)$ by Eq.~(\ref{eqn:competence}); \\
		Uniformly sample a data batch, $B(0)$, from the top $c(0)$ portions of $D^\text{train}_1$;
		\REPEAT
		\STATE Train the model with the $B(t)$;
		\STATE $t \leftarrow t + 1$;
		\STATE Estimate the model competence, $c(t)$, by Eq.~(\ref{eqn:competence}); \\
		Uniformly sample a data batch, $B(t)$, from the top $c(t)$ portions of $D^\text{train}_1$;
		\UNTIL Model converge.
	\end{algorithmic}
\end{algorithm}

\subsection{Single Difficulty-based Curriculum Learning}
\citet{Platanios2019Competence} proposed a competence-based and single difficulty-based curriculum learning framework (see Algorithm~\ref{alg:single}), which first sorts each instance in the training dataset $D^\text{train}$ according to a single difficulty metric $d$, and then defines the model competence $c(t) \in (0,1]$ at training step $t$ by following functional forms:
\begin{equation}
\footnotesize
\label{eqn:competence}
c(t)= \min \left(1, \sqrt[p]{t \frac{1-{c(0)}^{p}}{T}+{c(0)}^{p}}\right)
\end{equation}
where $c(0)$ is the initial competence and usually set to 0.01, $p$ is the coefficient to control the curriculum schedule and is usually set to 2, and $T$ is the duration of curriculum learning and determines the length of the curriculum.
In implementations, at training time step $t$, the top $c(t)$ portions of the sorted training dataset are selected to sample a training batch to train the model.
In this way, the model is able to gradually proceed from easy samples to more complex ones in training, resulting in first starting to utilize the simple and easy-written reports for training, and then attempting to utilize harder reports for training.

\begin{algorithm}[t]
\small
	\renewcommand{\algorithmicrequire}{\textbf{Input:}}
	\renewcommand{\algorithmicensure}{\textbf{Output:}}
	\caption{\small{Multiple Difficulty-based Curriculum Learning. The {\color{red} Red} colored text denotes the differences from Algorithm~\ref{alg:single}.} }
	\label{alg:multiple}
	\begin{algorithmic}[1]
		\REQUIRE The training set $D^\text{train}$, $i \in \{1,2,3,4\}$.
		\ENSURE A model with multiple difficulty-based curriculum learning.
		\STATE Compute four difficulties, $d_i$, for each training sample in $D^\text{train}$; 
		\STATE Sort $D^\text{train}$ based each difficulty of every sample, resulting in $D^\text{train}_i$ (i.e., $D^\text{train}_1$, $D^\text{train}_2$, $D^\text{train}_3$, $D^\text{train}_4$);
		\FOR{$i = 1, 2, 3, 4$}
		\STATE $t_i = 0$; Initialize the model competence from $i^{th}$ perspective, $c_i(0)$, by Eq.~(\ref{eqn:competence}); Uniformly sample a data batch, $B_i(0)$, from the top $c_i(0)$ portions of $D^\text{train}_i$;
		\STATE {\color{red} Compute the perplexity (PPL) on $B_i(0)$, PPL($B_i(0)$)};
		\ENDFOR
		\REPEAT
		\STATE {\color{red}$j = \underset{i}{\arg\max} (\text{PPL}(B_i(t_i)))$};
		\STATE Train the model with the $B_j(t_j)$;
		\STATE $t_j \leftarrow t_j + 1$;
		\STATE Estimate the model competence from $j^{th}$ perspective, $c_j(t_j)$, by Eq.~(\ref{eqn:competence}); Uniformly sample a data batch, $B_j(t_j)$, from the top $c_j(t_j)$ portions of $D^\text{train}_j$;
		\STATE {\color{red} Compute the perplexity (PPL) of model on $B_j(t_j)$, PPL($B_j(t_j)$)};
		\UNTIL Model converge.
	\end{algorithmic}
\end{algorithm}

\subsection{Multiple Difficulty-based Curriculum Learning}
\label{sec:multiple}
The training instances of medical report generation task are pairs of medical images and corresponding reports which is a multi-modal data. It's hard to estimate the difficulty with only one metric. 
In addition, the experimental results (see Table~\ref{tab:quantitative}) show that directly fusing multiple difficulty metrics as one ($d_1+d_2+d_3+d_4$) is obviously inappropriate, which is also verified in \citet{Platanios2019Competence}.
To this end, we extend the single difficulty-based curriculum learning into the multiple difficulty-based curriculum learning, where we provide the medical report generation models with four different difficulty metrics, i.e., $d_1, d_2, d_3, d_4$ (see Section~\ref{sec:difficulty}).

\begin{figure}[t]
\centering
\includegraphics[width=1\linewidth]{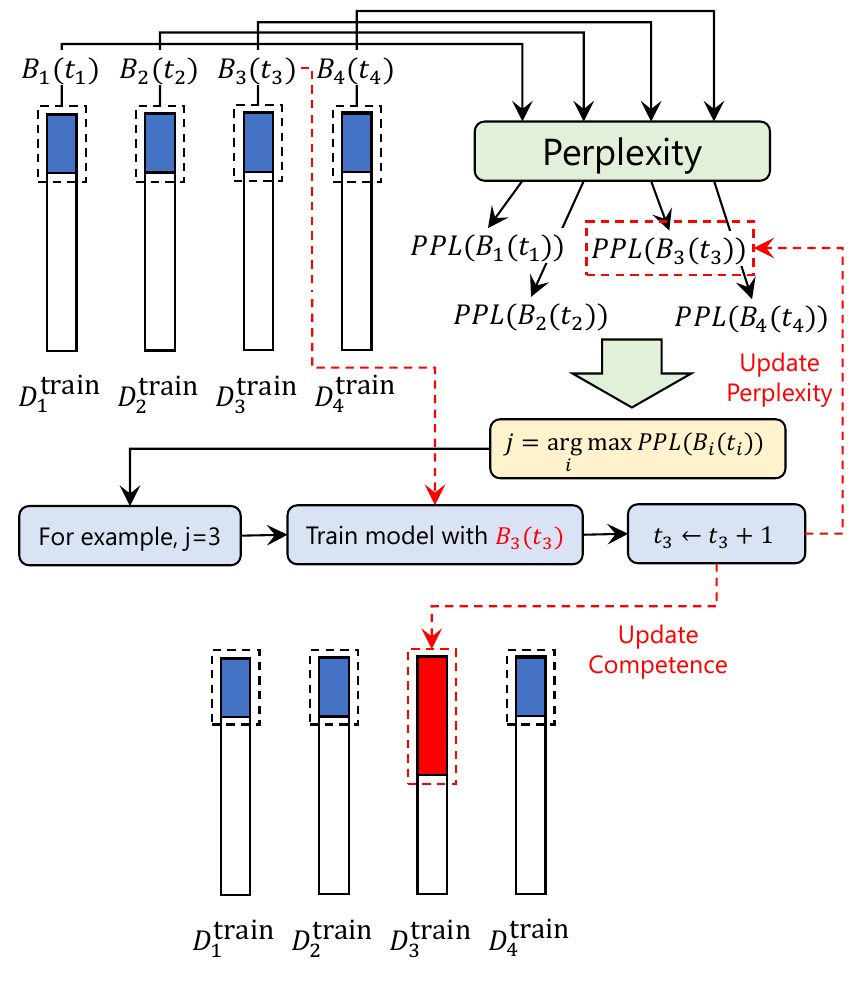}
\caption{Illustration of Algorithm~\ref{alg:multiple}.
}
\label{fig:algo}
\end{figure}

A simple and natural way is to randomly or sequentially choose a curricula to train the model, i.e., 1$\rightarrow$2$\rightarrow$3$\rightarrow$4$\rightarrow$1. However, a better approach is to adaptively select the most appropriate curricula for each training step, which follows the common practice of human learning behavior: 
When we have learned some curricula well, we tend to choose the under-learned curricula to learn.
Algorithm~\ref{alg:multiple} summarizes the overall learning process of the proposed framework and Figure \ref{fig:algo} illustrates the process of Algorithm~\ref{alg:multiple}.
In implementations, similarly, we first sort the training dataset based on the four difficulty metrics and acquire four sorted training datasets in line 1-2.
Then, based on the model competence, we acquire the training samples for each curricula, in line 4.
In line 5, we further estimate the perplexity (PPL) of model on different training samples $B_i(t_i)$ corresponding to different curricula, defined as:
\begin{equation}
\scriptsize
\label{eq:perplexity}
{\normalsize\text{PPL}}(B_i(t_i))=\sum_{R^{k} \in B_i(t_i)}\sqrt[N]{\prod_{m=1}^N \frac{1}{P(w^k_m|w^k_1,\dots,w^k_{m-1})}} \nonumber
\end{equation}
where $R^k=\{w^k_1,w^k_2,\dots,w^k_N\}$ denotes the $k$-th report in $B_i(t_i)$.
The perplexity (PPL) measures how many bits on average would be needed to encode each word of the report given the model, so the current curricula with higher PPL means that the model is not well-learned for this curricula and need to be improved.
Therefore, the PPL can be used to determine the curricula at each training step dynamically.
Specifically, in line 8-9, we select the under-learned curricula, i.e., the curricula with maximum PPL, to train the current model.
After that, we again estimate the model competence in the selected curricula in line 11 and compute the PPL of model on the training samples corresponding to the selected curricula in line 12.

\section{Experiment}
\label{sec:experiment}
We firstly describe two public datasets as well as the widely-used metrics, baselines and settings. Then we present the evaluation of our CMCL.

\subsection{Datasets}

We conduct experiments on two public datasets, i.e., a widely-used benchmark IU-Xray \cite{Dina2016IU-Xray} and a recently released large-scale MIMIC-CXR \cite{Johnson2019MIMIC}.

\begin{itemize}

    \item \textbf{IU-Xray}\footnote{\url{https://openi.nlm.nih.gov/}} is collected by Indiana University and is widely-used to evaluate the performance of medical report generation methods. It contains 7,470 chest X-ray images associated with 3,955 radiology reports sourced from Indiana Network for Patient Care.
 
    \item \textbf{MIMIC-CXR}\footnote{\url{https://physionet.org/content/mimic-cxr/2.0.0/}} is the recently released largest dataset to date and consists of 377,110 chest X-ray images and 227,835 radiology reports from 64,588 patients of the Beth Israel Deaconess Medical Center.
\end{itemize}
For IU-Xray dataset, following previous works \cite{Chen2020Generating,Jing2019Show,Li2019Knowledge,Li2018Hybrid}, we randomly split the dataset into 70\%-10\%-20\% training-validation-testing splits.
At last, we preprocess the reports by tokenizing, converting to lower-cases and removing non-alpha tokens.
For MIMIC-CXR, following \citet{Chen2020Generating}, we use the official splits to report our results, resulting in 368,960 samples in the training set, 2,991 samples in the validation set and 5,159 samples in the test set.
We convert all tokens of reports to lower-cases and filter tokens that occur less than 10 times in the corpus, resulting in a vocabulary of around 4,000 tokens.

\subsection{Baselines}
We tested three representative baselines that were originally designed for image captioning and three competitive baselines that were originally designed for medical report generation.

\subsubsection{Image Captioning Baselines}

\begin{itemize}
    \item \textbf{NIC:} \  
    \citet{Vinyals2015Show} proposed the encoder-decoder network, which employs a CNN-based encoder to extract image features and a RNN-based decoder to generate the target sentence, for image captioning. 

    \item \textbf{Spatial-Attention:} \ \citet{lu2017knowing} proposed the visual attention, which is calculated on the hidden states, to help the model to focus on the most relevant image regions instead of the whole image.
     
    \item \textbf{Adaptive-Attention:} \     
    Considering that the decoder tends to require little or no visual information from the image to predict the non-visual words such as ``the'' and ``of'', \citet{lu2017knowing} designed an adaptive attention model to decide when to employ the visual attention.

\end{itemize}

\subsubsection{Medical Report Generation Baselines}

\begin{itemize}
    \item \textbf{CNN-HLSTM:} \ \citet{Jing2018Automatic} introduced the Hierarchical LSTM structure (HLSTM), which contains the paragraph LSTM and the sentence LSTM. HLSTM first uses the paragraph LSTM to generate a series of high-level topic vectors representing the sentences, and then utilizes the sentence LSTM to generate a sentence based on each topic vector.

    \item \textbf{HLSTM+att+Dual:} \
    \citet{Harzig2019Addressing} proposed a hierarchical LSTM with the attention mechanism and further introduced two LSTMs, i.e., Normal LSTM and Abnormal LSTM, to help the model to generate more accurate normal and abnormal sentences.

    \item \textbf{Co-Attention:} \  \citet{Jing2018Automatic} proposed the co-attention model, which combines the merits of visual attention and semantic attention, to attend to both images and predicted semantic tags\footnote{\url{https://ii.nlm.nih.gov/MTI/}} simultaneously, exploring the synergistic effects of visual and semantic information.
    
\end{itemize}

\subsection{Metrics and Settings}
We adopt the widely-used BLEU \cite{papineni2002bleu}, METEOR \cite{Banerjee2005METEOR} and ROUGE-L \cite{lin2004rouge}, which are reported by the evaluation toolkit \cite{chen2015microsoft}\footnote{\url{https://github.com/tylin/coco-caption}}, to test the performance.
Specifically, ROUGE-L is proposed for automatic evaluation of the extracted text summarization. METEOR and BLEU are originally designed for machine translation evaluation.  

For all baselines, since our focus is to change the training paradigm, which improves existing baselines by efficiently utilizing the limited medical data, we keep the inner structure of the baselines untouched and preserve the original parameter setting.
For our curriculum learning framework, following previous work \cite{Platanios2019Competence}, the $c(0)$ and $p$ are set to 0.01 and 2, respectively.
For different baselines, we first re-implement the baselines without using any curriculum. When equipping baselines with curriculum, following \citet{Platanios2019Competence}, we set $T$ in Eq.(\ref{eqn:competence}) to a quarter of the number of training steps that the baseline model takes to reach approximately 90\% of its final BLEU-4 score.
To boost the performance, we further incorporate the Batching method  \cite{Xu2020Dynamic}, which batches the samples with similar difficulty in the curriculum learning framework.
To re-implement the baselines and our approach, following common practice \cite{Jing2019Show,Li2019Knowledge,Li2018Hybrid,fenglin2021PPKED,liu2021Contrastive}, we extract image features for both dataset used for evaluation from a ResNet-50 \cite{he2016deep}, which is pretrained on ImageNet \cite{Deng2009ImageNet} and fine-tuned on public available CheXpert dataset \cite{Irvin2019CheXpert}.
To ensure consistency with the experiment settings of previous works \cite{Chen2020Generating}, for IU-Xray, we utilize paired images of a patient as the input; for MIMIC-CXR, we use single image as the input.
For parameter optimization, we use Adam optimizer \cite{kingma2014adam} with a batch size of 16 and a learning rate of 1e-4.

\begin{table*}[t]
\centering
\scriptsize
\setlength{\tabcolsep}{5pt}
\begin{tabular}{cccccccccccccc} 
\shline
\multicolumn{1}{l|}{\multirow{2}{*}{Methods}}&  \multicolumn{6}{c|}{Dataset: MIMIC-CXR \cite{Johnson2019MIMIC}} &  \multicolumn{6}{c}{Dataset: IU-Xray \cite{Dina2016IU-Xray}}\\ \cline{2-13} 

\multicolumn{1}{c|}{} & \multicolumn{1}{c|}{B-1} &  \multicolumn{1}{c|}{B-2}         & \multicolumn{1}{c|}{B-3}         & \multicolumn{1}{c|}{B-4}         & \multicolumn{1}{c|}{M}         & \multicolumn{1}{c|}{R-L} &   \multicolumn{1}{c|}{B-1}&  \multicolumn{1}{c|}{B-2}         & \multicolumn{1}{c|}{B-3}         & \multicolumn{1}{c|}{B-4}         & \multicolumn{1}{c|}{M}         & \multicolumn{1}{c}{R-L}   \\ \hline \hline

\multicolumn{1}{l|}{NIC \cite{Vinyals2015Show}\ssymbol{2}} & \multicolumn{1}{c|}{0.290} & \multicolumn{1}{c|}{0.182} & \multicolumn{1}{c|}{0.119} & \multicolumn{1}{c|}{0.081} & \multicolumn{1}{c|}{0.112} & \multicolumn{1}{c|}{\bf 0.249} & \multicolumn{1}{c|}{0.352} & \multicolumn{1}{c|}{\bf 0.227} & \multicolumn{1}{c|}{0.154} & \multicolumn{1}{c|}{0.109} & \multicolumn{1}{c|}{0.133} & \multicolumn{1}{c }{0.313}  \\

\multicolumn{1}{l|}{ \ w/ CMCL } & \multicolumn{1}{c|}{\bf 0.301} & \multicolumn{1}{c|}{\bf 0.189} & \multicolumn{1}{c|}{\bf 0.123} & \multicolumn{1}{c|}{\bf 0.085} & \multicolumn{1}{c|}{\bf 0.119} & \multicolumn{1}{c|}{0.241} & \multicolumn{1}{c|}{\bf 0.358} & \multicolumn{1}{c|}{0.223} & \multicolumn{1}{c|}{\bf 0.160} & \multicolumn{1}{c|}{\bf 0.114} & \multicolumn{1}{c|}{\bf 0.137} & \multicolumn{1}{c}{\bf 0.317} \\ \hline \hline

\multicolumn{1}{l|}{Spatial-Attention \cite{lu2017knowing}\ssymbol{2}}  &\multicolumn{1}{c|}{0.302} & \multicolumn{1}{c|}{0.189} & \multicolumn{1}{c|}{0.122} & \multicolumn{1}{c|}{0.082} & \multicolumn{1}{c|}{\bf 0.120} & \multicolumn{1}{c|}{\bf 0.259} & \multicolumn{1}{c|}{0.374} & \multicolumn{1}{c|}{0.235} & \multicolumn{1}{c|}{0.158} & \multicolumn{1}{c|}{0.120} & \multicolumn{1}{c|}{0.146} & \multicolumn{1}{c}{0.322} \\

\multicolumn{1}{l|}{ \ w/ CMCL } & \multicolumn{1}{c|}{\bf 0.312} & \multicolumn{1}{c|}{\bf 0.200} & \multicolumn{1}{c|}{\bf 0.125} & \multicolumn{1}{c|}{\bf 0.087} & \multicolumn{1}{c|}{0.118} & \multicolumn{1}{c|}{0.258} & \multicolumn{1}{c|}{\bf 0.381} & \multicolumn{1}{c|}{\bf 0.246} & \multicolumn{1}{c|}{\bf 0.164} & \multicolumn{1}{c|}{\bf 0.123} & \multicolumn{1}{c|}{\bf 0.153} & \multicolumn{1}{c}{\bf 0.327} \\ \hline \hline

\multicolumn{1}{l|}{Adaptive-Attention \cite{lu2017knowing}\ssymbol{2}} &           \multicolumn{1}{c|}{\bf 0.307} & \multicolumn{1}{c|}{\bf 0.192} & \multicolumn{1}{c|}{0.124} & \multicolumn{1}{c|}{0.084} & \multicolumn{1}{c|}{0.119} & \multicolumn{1}{c|}{0.262} & \multicolumn{1}{c|}{0.433} & \multicolumn{1}{c|}{\bf 0.285} & \multicolumn{1}{c|}{0.194} & \multicolumn{1}{c|}{0.137} & \multicolumn{1}{c|}{0.166} & \multicolumn{1}{c }{\bf 0.349}\\

\multicolumn{1}{l|}{ \ w/ CMCL } & \multicolumn{1}{c|}{0.302} & \multicolumn{1}{c|}{\bf 0.192} & \multicolumn{1}{c|}{\bf 0.129} & \multicolumn{1}{c|}{\bf 0.091} & \multicolumn{1}{c|}{\bf 0.125} & \multicolumn{1}{c|}{\bf 0.264} & \multicolumn{1}{c|}{\bf 0.437} & \multicolumn{1}{c|}{0.281} & \multicolumn{1}{c|}{\bf 0.196} & \multicolumn{1}{c|}{\bf 0.140} & \multicolumn{1}{c|}{\bf 0.174} & \multicolumn{1}{c}{0.338}  \\ \hline \hline

\multicolumn{1}{l|}{CNN-HLSTM \cite{Krause2017Hierarchical}\ssymbol{2}} & \multicolumn{1}{c|}{0.321} & \multicolumn{1}{c|}{0.203} & \multicolumn{1}{c|}{0.129} & \multicolumn{1}{c|}{0.092} & \multicolumn{1}{c|}{0.125} & \multicolumn{1}{c|}{0.270} & \multicolumn{1}{c|}{0.435} & \multicolumn{1}{c|}{0.280} & \multicolumn{1}{c|}{0.187} & \multicolumn{1}{c|}{0.131} & \multicolumn{1}{c|}{0.173} & \multicolumn{1}{c}{0.346} \\

\multicolumn{1}{l|}{ \ w/ CMCL } & \multicolumn{1}{c|}{\bf 0.337} & \multicolumn{1}{c|}{\bf 0.210} & \multicolumn{1}{c|}{\bf 0.136} & \multicolumn{1}{c|}{\bf 0.097} & \multicolumn{1}{c|}{\bf 0.131} & \multicolumn{1}{c|}{\bf 0.274} & \multicolumn{1}{c|}{\bf 0.462} & \multicolumn{1}{c|}{\bf 0.293} & \multicolumn{1}{c|}{\bf 0.207} & \multicolumn{1}{c|}{\bf 0.155} & \multicolumn{1}{c|}{\bf 0.179} & \multicolumn{1}{c}{\bf 0.360}  \\
\hline \hline

\multicolumn{1}{l|}{HLSTM+att+Dual  \cite{Harzig2019Addressing}\ssymbol{2}}     & \multicolumn{1}{c|}{0.328} & \multicolumn{1}{c|}{0.204} & \multicolumn{1}{c|}{0.127} & \multicolumn{1}{c|}{\bf 0.090} & \multicolumn{1}{c|}{\bf 0.122} & \multicolumn{1}{c|}{0.267} & \multicolumn{1}{c|}{0.447} & \multicolumn{1}{c|}{0.289} & \multicolumn{1}{c|}{0.192} & \multicolumn{1}{c|}{0.144} & \multicolumn{1}{c|}{\bf 0.175} & \multicolumn{1}{c }{0.358}\\

\multicolumn{1}{l|}{ \ w/ CMCL } & \multicolumn{1}{c|}{\bf 0.330} & \multicolumn{1}{c|}{\bf 0.206} & \multicolumn{1}{c|}{\bf 0.133} & \multicolumn{1}{c|}{0.088} & \multicolumn{1}{c|}{0.119} & \multicolumn{1}{c|}{\bf 0.272} & \multicolumn{1}{c|}{\bf 0.461} & \multicolumn{1}{c|}{\bf 0.298} & \multicolumn{1}{c|}{\bf 0.201} & \multicolumn{1}{c|}{\bf 0.150} & \multicolumn{1}{c|}{0.173} & \multicolumn{1}{c}{\bf 0.359} \\
\hline \hline

\multicolumn{1}{l|}{Co-Attention \cite{Jing2018Automatic}\ssymbol{2}}& \multicolumn{1}{c|}{0.329} & \multicolumn{1}{c|}{0.206} & \multicolumn{1}{c|}{0.133} & \multicolumn{1}{c|}{0.095} & \multicolumn{1}{c|}{0.129} & \multicolumn{1}{c|}{0.273} & \multicolumn{1}{c|}{0.463} & \multicolumn{1}{c|}{0.293} & \multicolumn{1}{c|}{0.207} & \multicolumn{1}{c|}{0.155} & \multicolumn{1}{c|}{0.178} & \multicolumn{1}{c }{0.365} \\

\multicolumn{1}{l|}{ \ w/ CMCL } & \multicolumn{1}{c|}{\bf 0.344} & \multicolumn{1}{c|}{\bf 0.217} & \multicolumn{1}{c|}{\bf 0.140} & \multicolumn{1}{c|}{\bf 0.097} & \multicolumn{1}{c|}{\bf 0.133} & \multicolumn{1}{c|}{\bf 0.281} & \multicolumn{1}{c|}{\bf 0.473} & \multicolumn{1}{c|}{\bf 0.305} & \multicolumn{1}{c|}{\bf 0.217} & \multicolumn{1}{c|}{\bf 0.162} & \multicolumn{1}{c|}{\bf 0.186} & \multicolumn{1}{c}{\bf 0.378} \\
\hline \shline

\end{tabular}
\caption{Performance of automatic evaluations on the test sets of the MIMIC-CXR and the IU-Xray datasets. CMCL denotes the Competence-based Multimodal Curriculum Learning framework. B-n, M and R-L are short for BLEU-n, METEOR and ROUGE-L, respectively. Higher is better in all columns. \ssymbol{2} denotes our re-implementation. As we can see, all baseline models enjoy comfortable improvements in most metrics with our CMCL.
\label{tab:automatic}}
\end{table*}

\begin{table*}[t]
\centering
\scriptsize
\setlength{\tabcolsep}{5pt}
\begin{tabular}{cccccccccccccc} 
\shline
\multicolumn{1}{l|}{\multirow{2}{*}{Methods}}&  \multicolumn{6}{c|}{Dataset: MIMIC-CXR \cite{Johnson2019MIMIC}} &  \multicolumn{6}{c}{Dataset: IU-Xray \cite{Dina2016IU-Xray}}\\ \cline{2-13} 

\multicolumn{1}{c|}{} & \multicolumn{1}{c|}{B-1} &  \multicolumn{1}{c|}{B-2}         & \multicolumn{1}{c|}{B-3}         & \multicolumn{1}{c|}{B-4}         & \multicolumn{1}{c|}{M}         & \multicolumn{1}{c|}{R-L} &   \multicolumn{1}{c|}{B-1}&  \multicolumn{1}{c|}{B-2}         & \multicolumn{1}{c|}{B-3}         & \multicolumn{1}{c|}{B-4}         & \multicolumn{1}{c|}{M}         & \multicolumn{1}{c}{R-L}   \\ \hline \hline

\multicolumn{1}{l|}{HRGR-Agent \cite{Li2018Hybrid}} & \multicolumn{1}{c|}{-} & \multicolumn{1}{c|}{-} & \multicolumn{1}{c|}{-} & \multicolumn{1}{c|}{-} & \multicolumn{1}{c|}{-} & \multicolumn{1}{c|}{-} & \multicolumn{1}{c|}{0.438} & \multicolumn{1}{c|}{0.298} & \multicolumn{1}{c|}{0.208} & \multicolumn{1}{c|}{0.151} & \multicolumn{1}{c|}{-} & \multicolumn{1}{c}{0.322}  \\

\multicolumn{1}{l|}{CMAS-RL \cite{Jing2019Show}} & \multicolumn{1}{c|}{-} & \multicolumn{1}{c|}{-} & \multicolumn{1}{c|}{-} & \multicolumn{1}{c|}{-} & \multicolumn{1}{c|}{-} & \multicolumn{1}{c|}{-} & \multicolumn{1}{c|}{0.464} & \multicolumn{1}{c|}{0.301} & \multicolumn{1}{c|}{0.210} & \multicolumn{1}{c|}{0.154} & \multicolumn{1}{c|}{-} & \multicolumn{1}{c}{0.362}   \\

\multicolumn{1}{l|}{SentSAT + KG \cite{Zhang2020When}} & \multicolumn{1}{c|}{-} & \multicolumn{1}{c|}{-} & \multicolumn{1}{c|}{-} & \multicolumn{1}{c|}{-} & \multicolumn{1}{c|}{-} & \multicolumn{1}{c|}{-} & \multicolumn{1}{c|}{0.441} & \multicolumn{1}{c|}{0.291} & \multicolumn{1}{c|}{0.203} & \multicolumn{1}{c|}{0.147} & \multicolumn{1}{c|}{-} & \multicolumn{1}{c}{0.367}  \\

\multicolumn{1}{l|}{Up-Down \cite{anderson2018bottom}} & \multicolumn{1}{c|}{0.317} & \multicolumn{1}{c|}{0.195} & \multicolumn{1}{c|}{0.130} & \multicolumn{1}{c|}{0.092} & \multicolumn{1}{c|}{0.128} & \multicolumn{1}{c|}{0.267} & \multicolumn{1}{c|}{-}  & \multicolumn{1}{c|}{-} & \multicolumn{1}{c|}{-}  & \multicolumn{1}{c|}{-}  & \multicolumn{1}{c|}{-} & \multicolumn{1}{c}{-} \\

\multicolumn{1}{l|}{Transformer \cite{Chen2020Generating}} & \multicolumn{1}{c|}{0.314} & \multicolumn{1}{c|}{0.192} & \multicolumn{1}{c|}{0.127} & \multicolumn{1}{c|}{0.090} & \multicolumn{1}{c|}{0.125} & \multicolumn{1}{c|}{0.265} & \multicolumn{1}{c|}{0.396} &      \multicolumn{1}{c|}{0.254} & \multicolumn{1}{c|}{0.179} & \multicolumn{1}{c|}{0.135} & \multicolumn{1}{c|}{0.164} & \multicolumn{1}{c}{0.342} \\     

\multicolumn{1}{l|}{R2Gen \cite{Chen2020Generating}} & \multicolumn{1}{c|}{\color{red} 0.353} & \multicolumn{1}{c|}{\color{red} 0.218} & \multicolumn{1}{c|}{\color{red} 0.145} & \multicolumn{1}{c|}{\color{red} 0.103} & \multicolumn{1}{c|}{\color{red} 0.142} & \multicolumn{1}{c|}{\color{blue} 0.277} & \multicolumn{1}{c|}{\color{blue} 0.470} & \multicolumn{1}{c|}{\color{blue} 0.304} & \multicolumn{1}{c|}{\color{red} 0.219} & \multicolumn{1}{c|}{\color{red} 0.165} & \multicolumn{1}{c|}{\color{red} 0.187} & \multicolumn{1}{c}{\color{blue} 0.371} \\
\hline \hline

\multicolumn{1}{l|}{CMCL (Ours) } & \multicolumn{1}{c|}{\color{blue} 0.344} & \multicolumn{1}{c|}{\color{blue} 0.217} & \multicolumn{1}{c|}{\color{blue} 0.140} & \multicolumn{1}{c|}{\color{blue} 0.097} & \multicolumn{1}{c|}{\color{blue} 0.133} & \multicolumn{1}{c|}{\color{red} 0.281} & \multicolumn{1}{c|}{\color{red} 0.473} & \multicolumn{1}{c|}{\color{red} 0.305} & \multicolumn{1}{c|}{\color{blue} 0.217} & \multicolumn{1}{c|}{\color{blue} 0.162} & \multicolumn{1}{c|}{\color{blue} 0.186} & \multicolumn{1}{c}{\color{red} 0.378} \\

\hline \shline

\end{tabular}
\caption{Comparison with existing state-of-the-art methods on the test set of the MIMIC-CXR dataset and the IU-Xray dataset. CMCL is taken from the ``Co-Attention w/ CMCL'' in Table~\ref{tab:automatic}. In this table, the {\color{red} Red} and {\color{blue} Blue} colored numbers denote the best and second best results across all approaches, respectively.
\label{tab:automatic2}}
\end{table*}

\begin{table}[t]
\centering
\scriptsize

\setlength{\tabcolsep}{3pt}
\begin{tabular}{lccccccc} 
\shline
\multicolumn{1}{l|}{vs. Models} & \multicolumn{1}{c|}{Baseline wins} &  \multicolumn{1}{c|}{Tie}         & \multicolumn{1}{c}{`w/ CMCL' wins}         \\ \hline \hline

\multicolumn{1}{l|}{CNN-HLSTM \cite{Jing2018Automatic}\ssymbol{2}} & \multicolumn{1}{c|}{15} & \multicolumn{1}{c|}{28} & \multicolumn{1}{c}{\bf 57}   \\
\multicolumn{1}{l|}{Co-Attention \cite{Jing2018Automatic}\ssymbol{2}} & \multicolumn{1}{c|}{24} & \multicolumn{1}{c|}{35} & \multicolumn{1}{c}{\bf 41}  \\
\hline 
\shline
\end{tabular}
\caption{We invite 2 professional clinicians to conduct the human evaluation for comparing our method with baselines. All values are reported in percentage (\%).
\label{tab:human_evaluation}}
\end{table}

\begin{table*}[t]
\centering
\scriptsize

\begin{tabular}{ccccccccccccc}
\shline
\multicolumn{1}{c|}{\multirow{3}{*}{Settings}} & \multicolumn{2}{c|}{Visual Difficulty} & \multicolumn{2}{c|}{Textual Difficulty} & \multicolumn{1}{c|}{\multirow{3}{*}{\begin{tabular}[c]{@{}c@{}} Route \\ Strategy \end{tabular}}} 
&  \multicolumn{6}{c}{Dataset: IU-Xray \cite{Dina2016IU-Xray}}
\\ \cline{2-5} \cline{7-12}
\multicolumn{1}{c|}{} & \multicolumn{1}{c|}{\multirow{2}{*}{\begin{tabular}[c]{@{}c@{}} Heuristic \\ Metric \end{tabular}}} & \multicolumn{1}{c|}{\multirow{2}{*}{\begin{tabular}[c]{@{}c@{}} Model \\ Confidence \end{tabular}}} & \multicolumn{1}{c|}{\multirow{2}{*}{\begin{tabular}[c]{@{}c@{}} Heuristic \\ Metric \end{tabular}}} & \multicolumn{1}{c|}{\multirow{2}{*}{\begin{tabular}[c]{@{}c@{}} Model \\ Confidence \end{tabular}}} & \multicolumn{1}{c|}{} &  \multicolumn{6}{c}{Baseline: CNN-HLSTM \cite{Jing2018Automatic}}  \\  \cline{7-12}
\multicolumn{1}{c|}{} & \multicolumn{1}{c|}{} & \multicolumn{1}{c|}{} & \multicolumn{1}{c|}{} & \multicolumn{1}{c|}{} & \multicolumn{1}{c|}{}            & \multicolumn{1}{c|}{B-1}         &  \multicolumn{1}{c|}{B-2}         & \multicolumn{1}{c|}{B-3}         & \multicolumn{1}{c|}{B-4}         & \multicolumn{1}{c|}{M}           & \multicolumn{1}{c}{R-L}     \\ \hline \hline

\multicolumn{1}{c|}{Baseline} & \multicolumn{1}{c|}{-} & \multicolumn{1}{c|}{-} &\multicolumn{1}{l|}{-} & \multicolumn{1}{c|}{-} & \multicolumn{1}{c|}{-} & \multicolumn{1}{c|}{0.435} & \multicolumn{1}{c|}{0.280} & \multicolumn{1}{c|}{0.187} & \multicolumn{1}{c|}{0.131} & \multicolumn{1}{c|}{0.173} & \multicolumn{1}{c}{0.346}  \\ \hline \hline

\multicolumn{1}{c|}{(a)} &\multicolumn{1}{c|}{$\surd$} & \multicolumn{1}{c|}{-} & \multicolumn{1}{c|}{-} & \multicolumn{1}{c|}{-} & \multicolumn{1}{c|}{-} & \multicolumn{1}{c|}{0.438} & \multicolumn{1}{c|}{0.283} & \multicolumn{1}{c|}{0.188} & \multicolumn{1}{c|}{0.132} & \multicolumn{1}{c|}{0.173} & \multicolumn{1}{c}{0.348} \\

\multicolumn{1}{c|}{(b)} &\multicolumn{1}{c|}{-} & \multicolumn{1}{c|}{$\surd$} & \multicolumn{1}{c|}{-} & \multicolumn{1}{c|}{-} & \multicolumn{1}{c|}{-} & \multicolumn{1}{c|}{0.447} & \multicolumn{1}{c|}{0.288} & \multicolumn{1}{c|}{0.195} & \multicolumn{1}{c|}{0.143} & \multicolumn{1}{c|}{0.175} & \multicolumn{1}{c}{0.354} \\

\multicolumn{1}{c|}{(c)} &\multicolumn{1}{c|}{-} & \multicolumn{1}{c|}{-} & \multicolumn{1}{c|}{$\surd$} & \multicolumn{1}{c|}{-} & \multicolumn{1}{c|}{-} & \multicolumn{1}{c|}{0.443} & \multicolumn{1}{c|}{0.287} & \multicolumn{1}{c|}{0.192} & \multicolumn{1}{c|}{0.135} & \multicolumn{1}{c|}{0.175} & \multicolumn{1}{c}{0.351} \\

\multicolumn{1}{c|}{(d)} &\multicolumn{1}{c|}{-} & \multicolumn{1}{c|}{-} & \multicolumn{1}{c|}{-} & \multicolumn{1}{c|}{$\surd$} & \multicolumn{1}{c|}{-} & \multicolumn{1}{c|}{0.454} & \multicolumn{1}{c|}{0.290} & \multicolumn{1}{c|}{0.201} & \multicolumn{1}{c|}{0.148} & \multicolumn{1}{c|}{0.177} & \multicolumn{1}{c}{0.357} \\ \hline \hline

\multicolumn{1}{c|}{(e)} &\multicolumn{1}{c|}{$\surd$} & \multicolumn{1}{c|}{$\surd$} & \multicolumn{1}{c|}{-} & \multicolumn{1}{c|}{-} & \multicolumn{1}{c|}{Dynamically} & \multicolumn{1}{c|}{0.450} & \multicolumn{1}{c|}{0.289} & \multicolumn{1}{c|}{0.196} & \multicolumn{1}{c|}{0.144} & \multicolumn{1}{c|}{0.176} & \multicolumn{1}{c}{0.355}   \\

\multicolumn{1}{c|}{(f)} &\multicolumn{1}{c|}{$\surd$} & \multicolumn{1}{c|}{$\surd$} & \multicolumn{1}{c|}{$\surd$} & \multicolumn{1}{c|}{-} & \multicolumn{1}{c|}{Dynamically} & \multicolumn{1}{c|}{0.455} & \multicolumn{1}{c|}{0.290} & \multicolumn{1}{c|}{0.199} & \multicolumn{1}{c|}{0.145} & \multicolumn{1}{c|}{0.176} & \multicolumn{1}{c}{0.357} \\

\multicolumn{1}{c|}{\textbf{(g)}} &\multicolumn{1}{c|}{$\surd$} & \multicolumn{1}{c|}{$\surd$} & \multicolumn{1}{c|}{$\surd$} & \multicolumn{1}{c|}{$\surd$} & \multicolumn{1}{c|}{Dynamically} & \multicolumn{1}{c|}{\bf 0.462} & \multicolumn{1}{c|}{\bf 0.293} & \multicolumn{1}{c|}{\bf 0.207} & \multicolumn{1}{c|}{\bf 0.155} & \multicolumn{1}{c|}{\bf 0.179} & \multicolumn{1}{c}{\bf 0.360}  \\ \hline \hline

\multicolumn{1}{c|}{(h)} &\multicolumn{1}{c|}{$\surd$} & \multicolumn{1}{c|}{$\surd$} & \multicolumn{1}{c|}{$\surd$} & \multicolumn{1}{c|}{$\surd$} & \multicolumn{1}{c|}{Fuse} & \multicolumn{1}{c|}{0.440} & \multicolumn{1}{c|}{0.282} & \multicolumn{1}{c|}{0.190} & \multicolumn{1}{c|}{0.134} & \multicolumn{1}{c|}{0.174} & \multicolumn{1}{c}{0.349} \\

\multicolumn{1}{c|}{(i)} &\multicolumn{1}{c|}{$\surd$} & \multicolumn{1}{c|}{$\surd$} & \multicolumn{1}{c|}{$\surd$} & \multicolumn{1}{c|}{$\surd$} & \multicolumn{1}{c|}{Randomly} & \multicolumn{1}{c|}{0.457} & \multicolumn{1}{c|}{0.291} & \multicolumn{1}{c|}{0.199} & \multicolumn{1}{c|}{0.146} & \multicolumn{1}{c|}{0.178} & \multicolumn{1}{c}{0.358} \\

\multicolumn{1}{c|}{(j)} &\multicolumn{1}{c|}{$\surd$} & \multicolumn{1}{c|}{$\surd$} & \multicolumn{1}{c|}{$\surd$} & \multicolumn{1}{c|}{$\surd$} & \multicolumn{1}{c|}{Sequentially} & \multicolumn{1}{c|}{0.459} & \multicolumn{1}{c|}{0.290} & \multicolumn{1}{c|}{0.203} & \multicolumn{1}{c|}{0.150} & \multicolumn{1}{c|}{0.176} & \multicolumn{1}{c}{0.354} \\ 

\hline 
\shline
\end{tabular}
\caption{Quantitative analysis of our approach, which includes four designed difficulty metrics (see Section~\ref{sec:difficulty}) and the route strategy (see Section~\ref{sec:multiple}). We conduct the analysis on the widely-used baseline model CNN-HLSTM \cite{Jing2018Automatic}. The setting (g) also denotes our full proposed approach.}
\label{tab:quantitative}
\end{table*}

\subsection{Automatic Evaluation}
As shown in Table~\ref{tab:automatic}, for two datasets, all baselines equipped with our approach receive performance gains over most metrics.
The results prove the effectiveness and the compatibility of our CMCL in promoting the performance of existing models by better utilizing the limited medical data.
Besides, in Table~\ref{tab:automatic2}, we further select six existing state-of-the-art models, i.e., HRGR-Agent \cite{Li2018Hybrid}, CMAS-RL \cite{Jing2019Show}, SentSAT + KG \cite{Zhang2020When}, Up-Down \cite{anderson2018bottom}, Transformer \cite{Chen2020Generating} and R2Gen \cite{Chen2020Generating}, for comparison. For these selected models, we directly quote the results from the original paper for IU-Xray, and from \citet{Chen2020Generating} for MIMIC-CXR.
As we can see, based on the Co-Attention \cite{Chen2020Generating}, our approach CMCL achieves results competitive with these state-of-the-art models on major metrics, which further demonstrate the effectiveness of the proposed approach.

\subsection{Human Evaluation}

In this section, to verify the effectiveness of our approach in clinical practice, we invite two professional clinicians to evaluate the perceptual quality of 100 randomly selected reports generated by ``Baselines'' and ``Baselines w/ CMCL''.
For the baselines, we choose a representative model: CNN-HLSTM and a state-of-the-art model: Co-Attention.
The clinicians are unaware of which model generates these reports.
In particular, to have more documents examined, we did not use the same documents for both clinicians and check the agreements between them. That is to say, the documents for different clinicians do not overlap.
The results in Table~\ref{tab:human_evaluation} show that our approach is better than baselines in clinical practice with winning pick-up percentages.
In particular, all invited professional clinicians found that our approach can generate fluent reports with more accurate descriptions of abnormalities than baselines.
It indicates that our approach can help baselines to efficiently alleviate the data bias problem, which also can be verified in Section~\ref{sec:qualitative}.

\subsection{Quantitative Analysis}
\paragraph{Analysis on the Difficulty Metrics} 
In this section, we conduct an ablation study by only using a single difficulty metric during the curriculum learning, i.e., single difficulty-based curriculum learning, to investigate the contribution of each difficulty metric in our framework and the results are shown in Table~\ref{tab:quantitative}.
Settings (a-d) show that every difficulty metric can boost the performance of baselines, which verify the effectiveness of our designed difficulty metrics.
In particular, 1) the model confidence in both visual and textual difficulties achieves better performance than the heuristic metrics. It shows that the model confidence is the more critical in neural models. 2) Both the model confidence and heuristic metrics in the textual difficulty achieve better performance than their counterparts in the visual difficulty, which indicates that the textual data bias is the more critical in textual report generation task. 
When progressively incorporate each difficulty metric, the performance will increase continuously (see settings (e-g)), showing that integrating different difficulty metrics can bring the improvements from different aspects, and the advantages of all difficulty metrics can be united as an overall improvement.

\paragraph{Analysis on the Route Strategy}
As stated in Section~\ref{sec:multiple}, to implement the multiple difficulty-based curriculum learning, three simple and natural ways is to: 1) \textbf{Fuse} multiple difficulty metrics directly as a single mixed difficulty metric, $d_1+d_2+d_3+d_4$; 2) \textbf{Randomly} choose a curricula and 3) \textbf{Sequentially} choose a curricula (i.e., 1$\rightarrow$2$\rightarrow$3$\rightarrow$4$\rightarrow$1) to train the model.
Table~\ref{tab:quantitative} (h-j) show the results of the three implementations.
As we can see, all route strategies are viable in practice with improved performance of medical report generation, which proves the effectiveness and robustness of our CMCL framework.
Besides, all of them perform worse than our approach (Setting (g)), which confirms the effectiveness of dynamically learning strategy at each training step.

\subsection{Qualitative Analysis}
\label{sec:qualitative}

In Figure~\ref{fig:introduction}, we give two intuitive examples to better understand our approach.
As we can see, our approach generates structured and robust reports, which show significant alignment with ground truth reports and are supported by accurate abnormal descriptions.
For example, the generated report correctly describes ``\textit{Blunting of right costophrenic}'' in the first example and ``\textit{Scoliosis is present}'' in the second example.
The results prove our arguments and verify the effectiveness of our proposed CMCL in alleviating the data bias problem by enabling the model to gradually proceed from easy to more complex instances in training.

\section{Conclusion}
\label{sec:conclusion}
In this paper, we propose the novel competence-based multimodal curriculum learning framework (CMCL) to alleviate the data bias by efficiently utilizing the limited medical data for medical report generation.
To this end, considering the difficulty of accurately capturing and describing the abnormalities, we first assess four sample difficulties of training data from the visual complexity and the textual complexity, resulting in four different curricula.
Next, CMCL enables the model to be trained with the appropriate curricula and gradually proceed from easy samples to more complex ones in training.
Experimental results demonstrate the effectiveness and the generalization capabilities of CMCL, which consistently boosts the performance of the baselines under most metrics.

\section*{Acknowledgments}
This work is partly supported by Tencent Medical AI Lab, Beijing, China. 
We would like to sincerely thank the clinicians Xiaoxia Xie, Jing Zhang and Minghui Shao of the Harbin Chest Hospital in China for providing the human evaluation.
We sincerely thank all the anonymous reviewers for their constructive comments and suggestions.
Xian Wu is the corresponding author of this paper.

\section*{Ethical Considerations}
In this work, we focus on helping a wide range of existing medical report generation systems alleviate the data bias by efficiently utilizing the limited medical data for medical report generation.
Our work can enable the existing systems to gradually proceed from easy samples to more complex ones in training, which is similar to the learning curve of radiologist: (1) first start from simple and easy-written reports; (2) and then attempt to consume harder reports, which consist of rare and diverse abnormalities.
As a result, our work can promote the usefulness of existing medical report generation systems in better \textbf{assisting} radiologists in clinical decision-makings and reducing their workload.
In particular, for radiologists, given a large amount of medical images, the systems can automatically generate medical reports, the radiologists only need to make revisions rather than write a new report from scratch.
We conduct the experiments on the public MIMIC-CXR and IU-Xray datasets.
All protected health information was de-identified.
De-identification was performed in compliance with Health Insurance Portability and Accountability Act (HIPAA) standards in order to facilitate public access to the datasets.
Deletion of protected health information (PHI) from structured data sources (e.g., database fields that provide patient name or date of birth) was straightforward. 
All necessary patient/participant consent has been obtained and the appropriate institutional forms have been archived.

\bibliographystyle{acl_natbib}
\bibliography{acl2021}

\end{document}